\documentclass[lettersize,journal]{IEEEtran}
\usepackage{amsmath,amsfonts}
\usepackage{algorithmic}
\usepackage{algorithm}
\usepackage{array}
\usepackage{textcomp}
\usepackage{stfloats}

\usepackage{url}
\usepackage{hyperref}
\hypersetup{colorlinks=true, linkcolor=black} 

\usepackage{verbatim}
\usepackage{graphicx}
\usepackage{caption}
\usepackage{subcaption}
\usepackage{cite}
\usepackage{balance}
\usepackage{multirow}
\usepackage{underscore}
\usepackage{booktabs}
\usepackage{soul}
\begin{document}

\title{LVI-GS: Tightly-coupled LiDAR-Visual-Inertial SLAM using 3D Gaussian Splatting}

\author{Huibin~Zhao$^{*}$, Weipeng~Guan$^{*}$,   Peng~Lu  
    \thanks{
        *Equal contribution; 
    
        The authors are with the Adaptive Robotic Controls Lab (ArcLab), Department of Mechanical Engineering, The University of Hong Kong, Hong Kong SAR, China.
        (corresponding author: lupeng@hku.hk). 
        }%
    \thanks{
        This work was supported by General Research Fund under Grant 17204222, and in part by the Seed Fund for Collaborative Research and General Funding Scheme-HKU-TCL Joint Research Center for Artificial Intelligence.
        }%
}


\maketitle

\begin{abstract}

3D Gaussian Splatting (3DGS) has shown its ability in rapid rendering and high-fidelity mapping.
In this paper, we introduce LVI-GS, a tightly-coupled LiDAR-Visual-Inertial mapping framework with 3DGS, which leverages the complementary characteristics of LiDAR and image sensors to capture both geometric structures and visual details of 3D scenes.
To this end, the 3D Gaussians are initialized from colourized LiDAR points and optimized using differentiable rendering.
In order to achieve high-fidelity mapping, we introduce a pyramid-based training approach to effectively learn multi-level features and incorporate depth loss derived from LiDAR measurements to improve geometric feature perception.
Through well-designed strategies for Gaussian-Map expansion, keyframe selection, thread management, and custom CUDA acceleration, our framework achieves real-time photo-realistic mapping.
Numerical experiments are performed to evaluate the superior performance of our method compared to state-of-the-art 3D reconstruction systems.
Videos of the evaluations can be found on our website: \url{https://kwanwaipang.github.io/LVI-GS/}.
\end{abstract}


\section{Introduction}
\IEEEPARstart{S}{imultaneous} Localization and Mapping (SLAM) systems are indispensable across various domains, including robotics, augmented reality, and autonomous navigation\cite{kazerouni2022survey}. These systems enable devices to understand and navigate complex environments by constructing maps while simultaneously estimating their own positions within these spaces. For effective SLAM, both accurate localization and comprehensive scene reconstruction are essential.

Traditional SLAM systems represent environments using landmark~\cite{huang2022vwr}, point clouds\cite{li2022intensity}, occupancy grids\cite{elfes1989using}, Signed Distance Function (SDF) voxel grids\cite{reijgwart2019voxgraph}, or meshes\cite{lin2023immesh}. Among these, point clouds are a straightforward scene representation readily obtainable from sensors like cameras and LiDAR. Point cloud-based SLAM systems can achieve accurate localization and construct either sparse or dense maps, though these tend to lack visually rich details.

The advent of Neural Radiance Fields (NeRF) has introduced a new approach for high-fidelity scene reconstruction\cite{mildenhall2021nerf}. NeRF implicitly represents a scene within a radiance field by optimizing a continuous volumetric scene function, which requires minimal memory. Some NeRF-based SLAM methods leverage this framework’s novel view synthesis and high-fidelity reconstruction capabilities to model scenes. For instance, iMAP\cite{imap} constructs an implicit 3D occupancy and colour model usable for tracking, while NICE-SLAM\cite{niceslam} represents larger scenes via a coarse-to-fine approach. Enhanced methods such as Vox-Fusion\cite{voxfusion}, CoSLAM\cite{coslam}, and ES-SLAM\cite{esslam} advance SLAM system performance to varying extents. However, due to the extensive optimization processes involved, these systems struggle to achieve real-time performance. Moreover, storing maps within multi-layer perceptrons poses challenges, including catastrophic forgetting and limited boundaries, which can impede scene reconstruction.

3D Gaussian Splatting (3DGS) offers an exciting alternative, providing a continuous and adaptable representation for modelling 3D scenes through differentiable 3D Gaussian-shaped primitives\cite{fei20243d}\cite{kerbl20233d}. As a semi-implicit mapping approach, it trades off some novel view synthesis capabilities for significantly faster optimization and rendering speeds. Despite being based on optimization, 3DGS closely resembles point and surfel clouds, thus inheriting their efficiency, locality, and adaptability—attributes beneficial for SLAM mapping. With rendering speeds up to 200 frames per second at a 1080p resolution, 3DGS also initializes using point clouds, allowing it to leverage the sparse or dense point clouds generated by conventional SLAM systems for high-fidelity imagery\cite{monogs}.

Recently, several SLAM approaches integrating 3D Gaussians have shown promising results. Methods such as SplaTAM\cite{splatam}, MonoGS\cite{monogs}, GS-SLAM\cite{gsslam}, and Photo-SLAM\cite{photoslam} employ sequential RGB-D or RGB data to establish complete SLAM systems. However, these techniques encounter difficulties in large-scale, uncontrolled outdoor environments characterized by challenging lighting, complex backgrounds, and rapid motion. Although LiDAR offers high-quality geometric initialization for 3D Gaussians and is generally more robust in outdoor settings than cameras, integrating it into SLAM systems introduces unique challenges. LIV-Gaussianmap\cite{livgaussmap} and LetsGo\cite{letsgo} utilize LiDAR for initializing 3D Gaussians, while Gaussian-LIC\cite{gaussianlic} combines a LiDAR-Inertial-Camera setup for comprehensive 3D Gaussian construction. Nevertheless, systems like LIV-Gaussianmap\cite{livgaussmap} and LetsGo\cite{letsgo} are limited to offline processing, and Gaussian-LIC\cite{gaussianlic} requires complex front-end odometry and maintaining a substantial number of keyframes.

In general, the main contributions of this study can be outlined as follows:
\begin{enumerate}
\item 
We have developed and implemented a sophisticated real-time LVI-GS system that is capable of maintaining a dynamic hyper primitives module. This system leverages 3D Gaussian Splatting (3DGS) to perform high-quality, real-time rendering in three-dimensional space, ensuring an efficient and accurate representation of complex environments.
\item 
To further enhance the performance and scalability of our system, we employed a coarse-to-fine map construction approach. This method utilizes both RGB image pyramids and depth image pyramids to progressively refine the map at different levels of detail. Additionally, we implemented an advanced thread management technique to optimize computational efficiency, ensuring smooth real-time operations even with large datasets.
\item 
In pursuit of improved map representation and rendering quality, we designed a robust keyframe management strategy that allows for the effective selection and processing of keyframes. Moreover, by incorporating depth loss into the system, we enhanced the accuracy of the 3D Gaussian map, leading to more precise reconstructions and visually superior rendering results.
\end{enumerate}

\section{RELATED WORKS}
\subsection{NeRF-based SLAM}
The classic NeRF-based SLAM systems entirely rely on MLP. They encode the environment within the weights of a neural network. By querying this function at different points along rays cast through the scene, it can render novel views with high visual fidelity and realistic lighting effects. As the first NeRF-based SLAM system, iMAP\cite{imap}, ensures close-to-frame-rate camera tracking by joint optimization of photometric and geometric losses. Nevertheless, constrained by the MLP's capacity, it results in less intricate reconstruction and encounters difficulties with catastrophic forgetting when dealing with larger environments. To address this issue, subsequent works focus on combining the strengths of implicit MLPs and structural features to significantly enhance scene scalability and precision. For example, NICE-SLAM\cite{niceslam} adopts a hierarchical strategy that integrates multi-level local data. This approach effectively addresses issues like excessively smoothed reconstructions and scalability limitations in larger scenes. Vox-Fusion uses octree to manage voxels, providing the scene with scalability. ESLAM\cite{esslam}, Co-SLAM\cite{coslam}, GO-SLAM\cite{goslam}, and Point-SLAM\cite{pointslam}, introduce innovative approaches such as multi-scale feature planes, combined smoothness, detail encoding, real-time global optimization, and dynamic neural point cloud representation, enhancing scene reconstruction, camera tracking, and map fidelity with improved memory efficiency and scalability. OrbeezSLAM\cite{orbeez}, NeRF-SLAM\cite{nerfslam} and NeRF-VO\cite{nerfvo} improve tracking frequency and pose estimation accuracy of Nerf-based SLAM methods for odometry.

\subsection{3DGS-based SLAM}
3DGS-based methods offer faster, more realistic rendering, enhanced map capacity, and efficient optimization through dense losses compared to NeRF-based SLAM. SplaTAM\cite{splatam} employs simplified 3D Gaussians for accurate camera pose estimation and scene refinement, enhancing rendering efficiency and mapping precision. In contrast, GS-SLAM\cite{gsslam} utilizes 3D Gaussians, opacity, and spherical harmonics for scene representation, featuring adaptive Gaussian management and robust camera tracking but facing challenges with depth data quality and memory usage in large scenes. MonoGS\cite{monogs} is the first work to achieve 3DGS SLAM using a monocular camera. GS-ICP\cite{gsicp} SLAM utilizes the explicit characteristics of 3D Gaussian Splatting by employing G-ICP\cite{gicp} for the tracking process. Photo-SLAM\cite{photoslam} utilizes accurate pose estimation from ORB-SLAM3\cite{orbslam3} to reconstruct a hybrid Gaussian map. Some methods incorporate LiDAR as a sensor. Gaussian-LIC\cite{gaussianlic} realize a real-time LiDAR-Inertial-Camera SLAM system relying on a tightly-coupled LiDAR-Inertial-Camera odometry. MM-Gaussian\cite{mmgaussian} develop a relocalization module that is designed to correct the system’s trajectory in the event of localization failures. Meanwhile, MM3DGS-SLAM\cite{mm3dgs} uses a visual-inertial framework to optimize camera poses. It addresses the scale ambiguity of monocular depth estimates.

\section{METHODOLOGY}
Our framework implements the complete system functionality through two parallel threads. One thread handles odometry, while the other performs real-time optimization of 3D Gaussians. Both threads collaboratively maintain a shared hyper primitives module. Between these two threads, data such as 3D point clouds, camera poses, camera images and depth information are exchanged.

\begin{figure*}[htb]  
        \captionsetup{justification=justified, labelsep=colon}
        \centering
        \includegraphics[width=2.0\columnwidth]{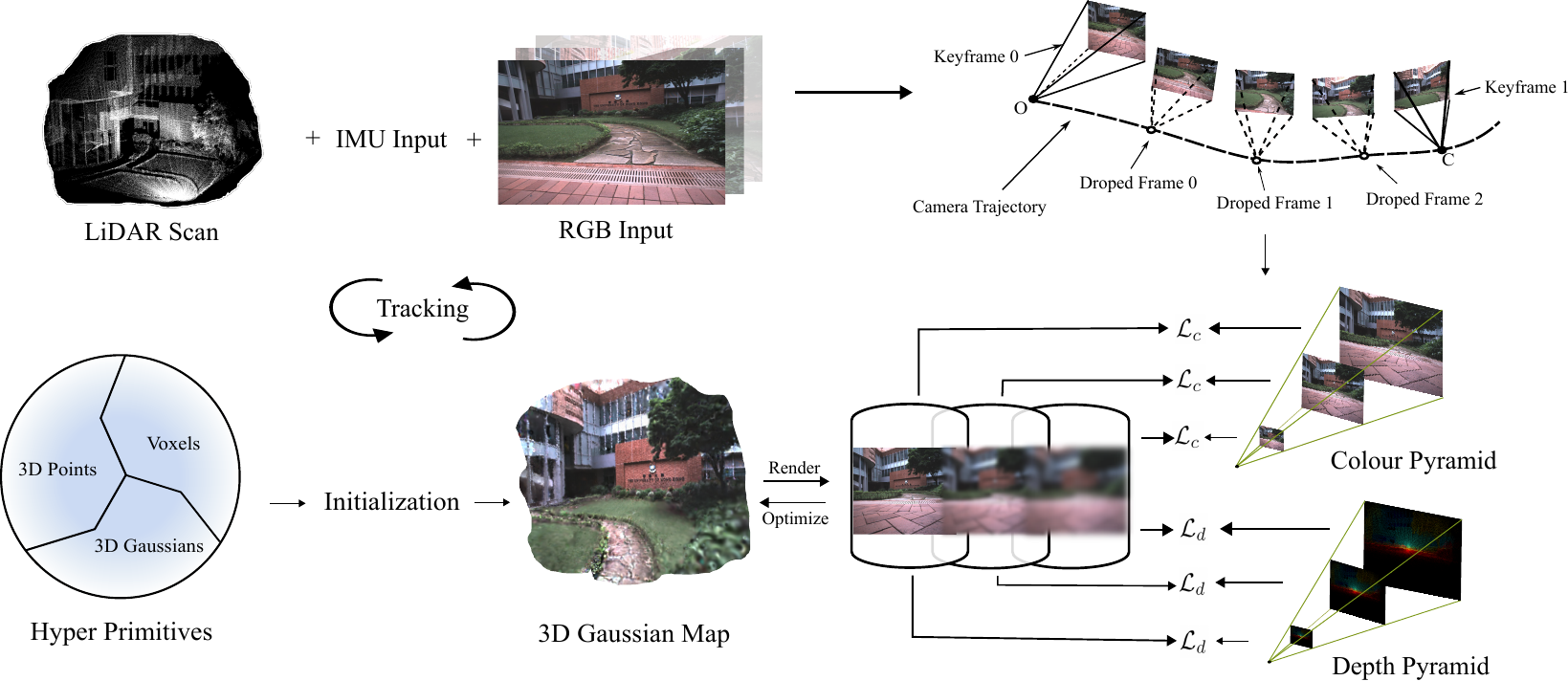}
        \caption{Overview of our proposed LVI-GS system.
        }  
        \label{pipeline}
\end{figure*}%

\subsection{Hyper Primitives}

We maintain a hyper primitives module, consisting of 3D point clouds, voxels, and 3D Gaussians. To efficiently access 3D point clouds for 3D Gaussian initialization, the map points are organized into fixed-size voxels (e.g., 0.1 m x 0.1 m x 0.1 m). Voxel activation is determined by the presence of recently added points (e.g., within the last second). An activated voxel indicates recent activity, whereas a deactivated voxel signifies no recent updates.

In addition, within the Visual-Inertial Odometry (VIO) module, points are removed if their projection or photometric errors exceed a specified threshold\cite{r3live}. For each point in the point cloud, we identify its position within the grid; if a point already exists in that position, it is discarded. We also regulate the number of points within each voxel to maintain a controlled density. Through this initial filtering process, as odometry proceeds, the obtained point cloud avoids the redundant addition of 3D Gaussians.

\subsection{3D Gaussian Splatting}
Our scene representation is 3DGS, mapping it with a set of anisotropic Gaussian $\mathcal{G}$. Each Gaussian includes opacity $o\in[0,1]$, center position $\mu\in\mathbb{R}^3$, RGB colour $c$, radius $r$ and 3D covariance matrix $\Sigma\in\mathbb{R}^{3\times3}$.
Given a center position $\mu$ and 3D covariance matrix $\Sigma$, a Gaussian distribution is defined as 
\begin{equation}
\mathcal{D}(x)= \exp(-\frac{1}{2}(x-\mu)\Sigma^{-1}(x-\mu)^{T})
\label{eq:Dx}
\end{equation}
Because every Gaussian shape is an ellipsoid, we parameterize the 3D Gaussian’s covariance as:
\begin{equation}
\Sigma = RSS^TR^T
\label{eq:Sigma}
\end{equation}
where $S\in\mathbb{R}^3$ is vector to describe 3D scale, $R\in\mathbb{R}^{3\times3}$ represent the rotation matrix. Instead of traversing along the camera rays, 3DGS conducts rasterization by iterating over the 3D Gaussians, thereby disregarding vacant spaces in the rendering process. Given 3DGS's utilization of volume rendering, there is no necessity for the direct derivation of the surface. Rather, through the amalgamation of $\mathcal{N}$ 3D Gaussians via splatting and blending, a pixel's colour $\mathcal{C}_{p}$ is determined:
\begin{equation}
\mathcal{C}_p=\sum_{i\in{\mathcal{N}}}{c}_{i}{\alpha}_{i}\prod_{j=1}^{i-1}(1-{\alpha}_{j})
\label{eq:Cp}
\end{equation}
Similarly, using the same method, we can also obtain the depth $\mathcal{D}_{p}$ through:
\begin{equation}
\mathcal{D}_p=\sum_{i\in{\mathcal{N}}}{d}_{i}{\alpha}_{i}\prod_{j=1}^{i-1}(1-{\alpha}_{j})
\label{eq:Dp}
\end{equation}
We also render a visibility image and it is used to determine the visibility of the current pixel.
\begin{equation}
\mathcal{V}_p=\sum_{i\in{\mathcal{N}}}{\alpha}_{i}\prod_{j=1}^{i-1}(1-{\alpha}_{j})
\label{eq:Vp}
\end{equation}
where the final opacity $\alpha_i$ is the multiplication result of the learned opacity $o_i$ and the Gaussian:
\begin{equation}
\mathcal{\alpha}_i= o_i\exp(-\frac{1}{2}(x^\prime-\mu^\prime)\Sigma^{\prime{-1}}(x^\prime-\mu^\prime)^{T})
\label{eq:alpha}
\end{equation}
where coordinate $x^\prime$ and $\mu^\prime$ are both in the projected space.

Our ultimate goal is to project 3D Gaussians onto a 2D plane for rendering to obtain high-fidelity images, a process commonly referred to as "splatting", When we obtain the sensor's position transformation $[{Q}_{iw},{T}_{iw}]$ between image and world. Thus, the function to project 3D Gaussians $(\mu_{W},\Sigma_{W})$ to 2D Gaussians $(\mu_{I},\Sigma_{I})$ is 
\begin{equation}
\mu_{I} =[{Q}_{iw},{T}_{iw}]\cdot\mu_{W}, \Sigma_{I} = J{Q}_{iw}\Sigma_{W}{Q}_{iw}^{T}J^{T} 
\label{eq:muI}
\end{equation}
where J is the linear approximation of the Jacobian matrix for the projection transformation. ${Q}_{iw}$ and ${T}_{iw}$ are respectively the rotational and translational components of the sensor's pose.

\subsection{Keyframe Management}

We obtain point clouds through the hyper primitives module, packaging every $N_l$ points along with the corresponding camera pose and image into a single frame for future selection. All colourized LiDAR points are utilized for initializing the 3D Gaussians. We conduct keyframe selection on the images, evaluating those free from motion blur. Images that exhibit rotation or translation beyond a specified threshold are added as keyframes to the keyframe sequence:

\begin{equation}
\left\lVert \mathbf{R}_i - \mathbf{R}_{i-1} \right\rVert > \tau_r \quad \text{or} \quad \left\lVert \mathbf{T}_i - \mathbf{T}_{i-1} \right\rVert > \tau_t
\label{eq:threshold}
\end{equation}

In addition, for each newly added keyframe, we evaluate its visual overlap with previous keyframes. If the overlap exceeds a specified threshold, indicating high similarity, the frame is deemed redundant and is not added to the sequence. Additionally, a filtering criterion is applied to Gaussian additions. We compute the cumulative opacity from the current keyframe's viewpoint, selecting 3D Gaussians that meet the transparency requirements. If $\mathcal{V}_p \leq \tau_{\alpha}$, then filter out the point. 

Moreover, before using all colourized LiDAR points from the keyframes for 3D Gaussian initialization, we incorporate a buffer container to delay the keyframe sequence integration into the map. This delay prevents the opacity of a Gaussian, initialized by a preceding frame, from quickly decreasing to a prunable threshold before it can be observed from a subsequent frame’s viewpoint, thereby ensuring that 3D Gaussians remain available for training from multiple perspectives.

\begin{table*}[htp]
    \centering
    \renewcommand{\arraystretch}{1.0} 
    \setlength{\tabcolsep}{12pt} 
    \caption{Quantitative performance comparison of different methods on sequences hku2(f0), LiDAR_Degenerate(f1), Visual_Challenge(f2), hku_campus_seq_00(r0), degenerate_seq_00(r1), degenerate_seq_01(r2).}
    \label{tab:quantitativecomparison1}
    \begin{tabular}{@{}llccccccc@{}} 
        \toprule
        \multirow{2}{*}{\textbf{Method}} & \multirow{2}{*}{\textbf{Metrics}} & \multicolumn{6}{c}{\textbf{View}} & \multirow{2}{*}{\textbf{Avg.}} \\
        \cmidrule(lr){3-8}
        & & f0 & f1 & f2 & r0 & r1 & r2 &  \\
        \midrule
        \multirow{3}{*}{NeRF-SLAM\cite{nerfslam}} & PSNR$\uparrow$ & 25.56 & 25.47 & 17.01 & 19.53 & 21.20 & 15.02 & 20.63 \\
        & SSIM$\uparrow$ & 0.629 & 0.702 & 0.513 & 0.645 & 0.534 & 0.711 & 0.622 \\
        & LPIPS$\downarrow$ & 0.281 & 0.272 & 0.512 & 0.455 & 0.364 & 0.328 & 0.369 \\
        \midrule
        \multirow{3}{*}{MonoGS\cite{monogs}} & PSNR$\uparrow$ & 23.58 & 23.45 & 17.04 & 16.19 & 15.03 & 15.03 & 18.39 \\
        & SSIM$\uparrow$ & 0.699 & 0.762 & 0.671 & 0.490 & 0.512 & 0.457 & 0.599 \\
        & LPIPS$\downarrow$ & 0.643 & 0.757 & 0.643 & 0.710 & 0.755 & 0.767 & 0.713 \\
        \midrule
        \multirow{3}{*}{SplaTAM\cite{splatam}} & PSNR$\uparrow$ & 25.51 & 27.40 & 17.84 & 17.10 & 19.24 & 18.30 & 20.90 \\
        & SSIM$\uparrow$ & 0.709 & 0.798 & 0.652 & 0.526 & 0.666 & 0.597 & 0.658 \\
        & LPIPS$\downarrow$ & 0.214 & 0.235 & 0.346 & 0.448 & 0.295 & 0.456 & 0.332 \\
        \midrule
        \multirow{3}{*}{Gaussian-LIC\cite{gaussianlic}} & PSNR$\uparrow$ & \textbf{29.89} & 31.28 & 23.90 & \textbf{25.27} & 22.47 & 23.49 & 26.05 \\
        & SSIM$\uparrow$ & \textbf{0.820} & 0.840 & \textbf{0.830} & \textbf{0.805} & 0.794 & 0.811 & 0.817 \\
        & LPIPS$\downarrow$ & \textbf{0.155} & 0.178 & \textbf{0.173} & \textbf{0.212} & 0.192 & 0.187 & 0.183 \\
        \midrule
        \multirow{3}{*}{Our LVI-GS} & PSNR$\uparrow$ & 27.90 & \textbf{33.90} & \textbf{25.38} & 24.47 & \textbf{22.60} & \textbf{23.84} & 26.35 \\
        & SSIM$\uparrow$ & 0.731 & \textbf{0.826} & \textbf{0.830} & 0.776 & \textbf{0.818} & \textbf{0.850} & 0.805 \\
        & LPIPS$\downarrow$ & 0.200 & \textbf{0.043} & 0.237 & 0.244 & \textbf{0.159} & \textbf{0.170} & 0.176 \\
        \bottomrule
    \end{tabular}
    \captionsetup{justification=centering, labelsep=colon} 
\end{table*}

\subsection{Pyramids-based Training}
In our large-scale 3D Gaussian scene representation, we adopt a progressive training approach to optimize the efficiency of training the 3D Gaussian fields while maintaining rendering quality. By utilizing colour and depth images at varying resolutions, we construct pyramids of colour and depth images, improving the training process by gradually refining the level of detail. Specifically, we partition the Gaussian map into multi-scale representations to capture different levels of detail. The input colour and depth images undergo repeated downsampling, allowing us to train the 3D Gaussian progressively from coarse to fine resolution. During training, we prioritize low-resolution data to optimize coarse-level detail. After a certain number of iterations, we progressively reduce the level of downsampling, eventually using the original input resolution to complete the training. This method ensures efficient training while preserving the high-quality representation of the 3D Gaussian scene at various levels of detail.

\begin{equation}
\begin{aligned}
l_0 & :  \left( I_r^n, D_r^n , \mathrm{P}^n (I_\text{gt}), \mathrm{P}^n (D_\text{gt}) \right) \\
l_1 & : \left( I_r^{n-1}, D_r^{n-1}, \mathrm{P}^{n-1} (I_\text{gt}),\mathrm{P}^{n-1} (D_\text{gt}) \right) \\
& \vdots \\
l_n & : \left( I_r^0, D_r^0,  \mathrm{P}^0 (I_\text{gt}), \mathrm{P}^0 (D_\text{gt}) \right)
\end{aligned}
\end{equation}

Here, $l$ denotes the level of the pyramid, $I_r$ represents the rendered colour image, and $D_r$ denotes the rendered depth image. $P{(I_\text{gt})}$ represents the colour pyramid, while $P{(D_\text{gt})}$ denotes the depth pyramid.

\begin{figure*}[ht]
    \centering
    \begin{subfigure}{0.24\textwidth}
        \centering
        \includegraphics[width=\linewidth]{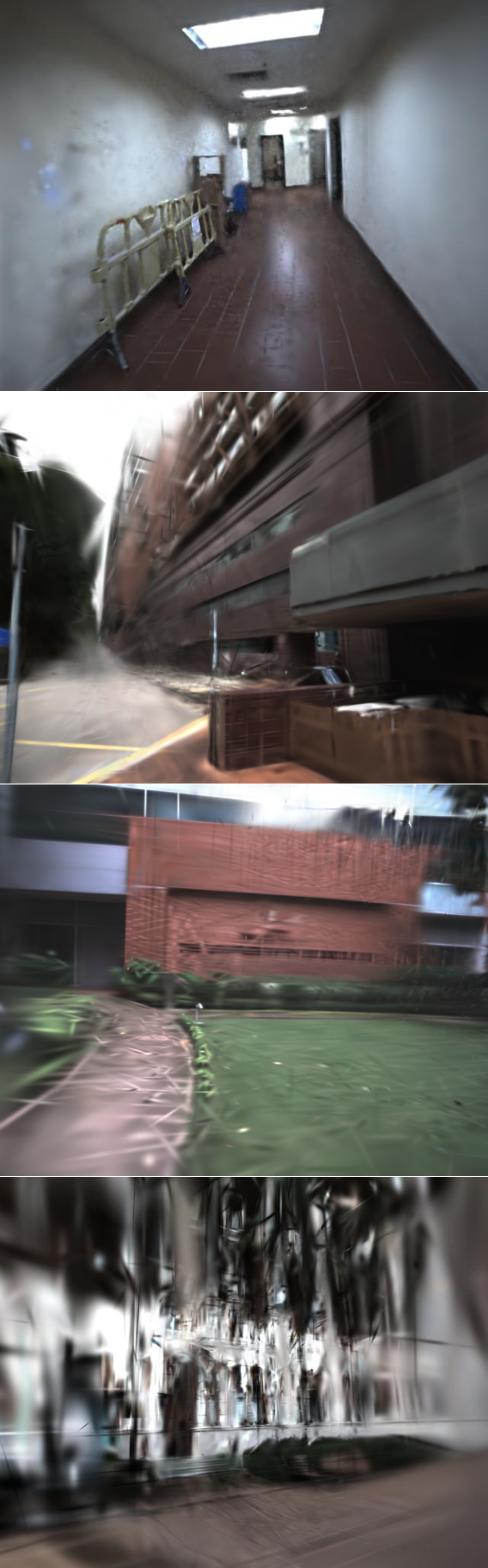}
        \caption{MonoGS\cite{monogs}}
        \label{fig:qualitativecomparisonmonogs}
    \end{subfigure}
    \hspace{0.0001\textwidth} 
    \begin{subfigure}{0.24\textwidth}
        \centering
        \includegraphics[width=\linewidth]{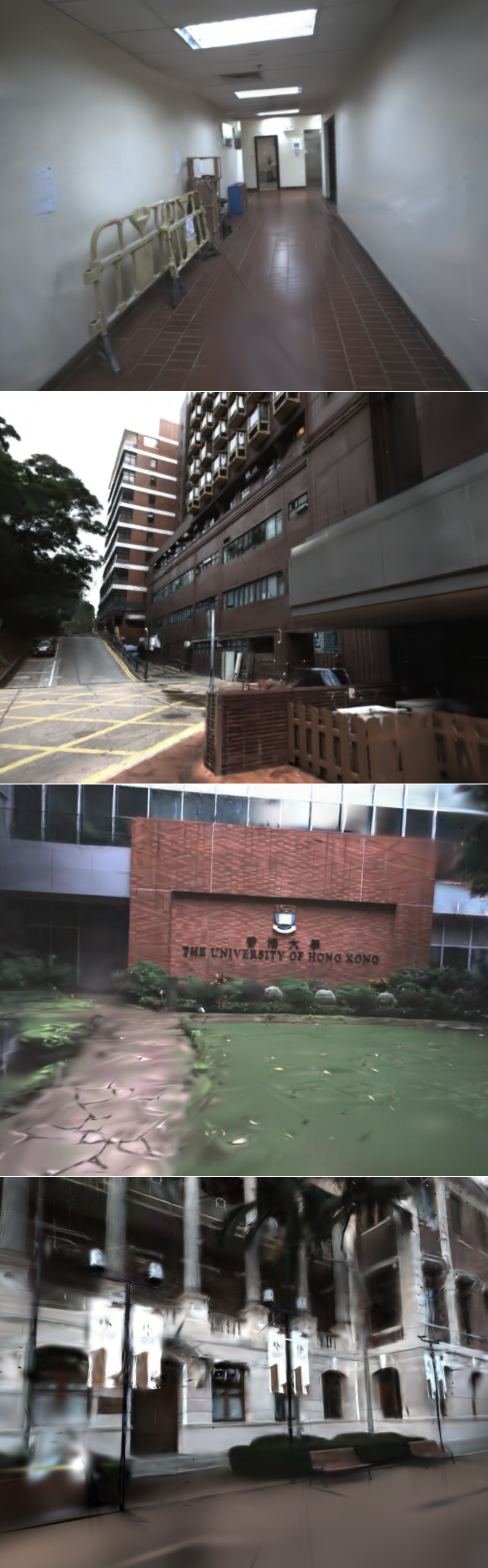}
        \caption{Photo-SLAM\cite{photoslam}}
        \label{fig:qualitativecomparisonphotoslam}
    \end{subfigure}
    \hspace{0.0001\textwidth} 
    \begin{subfigure}{0.24\textwidth}
        \centering
        \includegraphics[width=\linewidth]{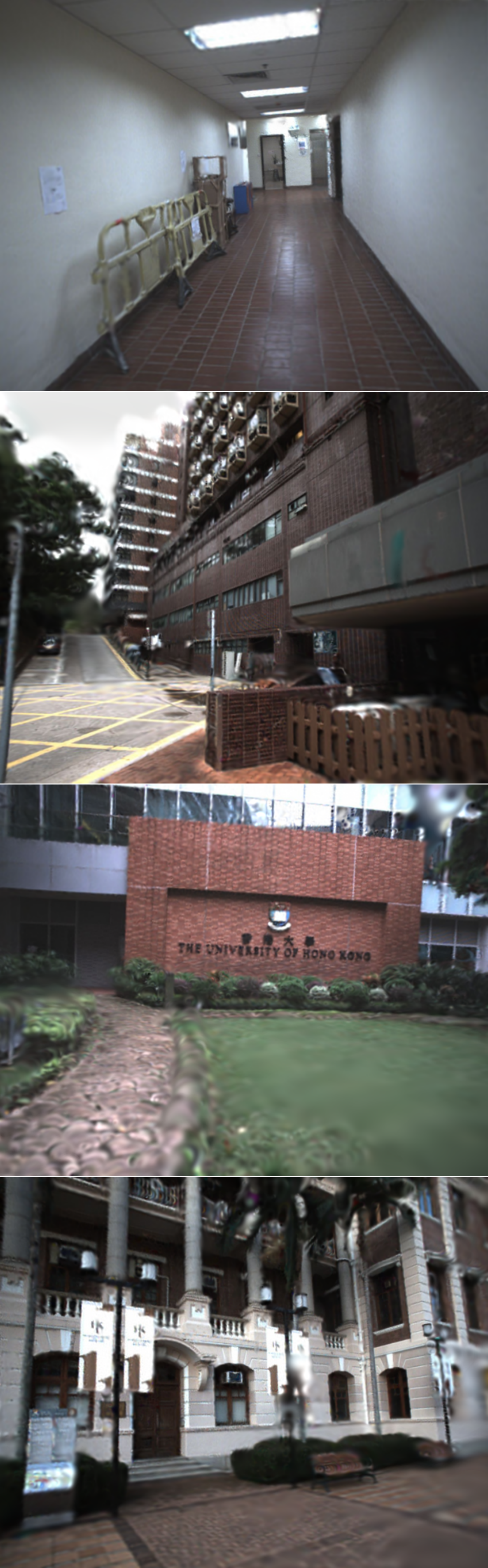}
        \caption{Ours}
        \label{fig:qualitativecomparisonours}
    \end{subfigure}
    \hspace{0.0001\textwidth} 
    \begin{subfigure}{0.24\textwidth}
        \centering
        \includegraphics[width=\linewidth]{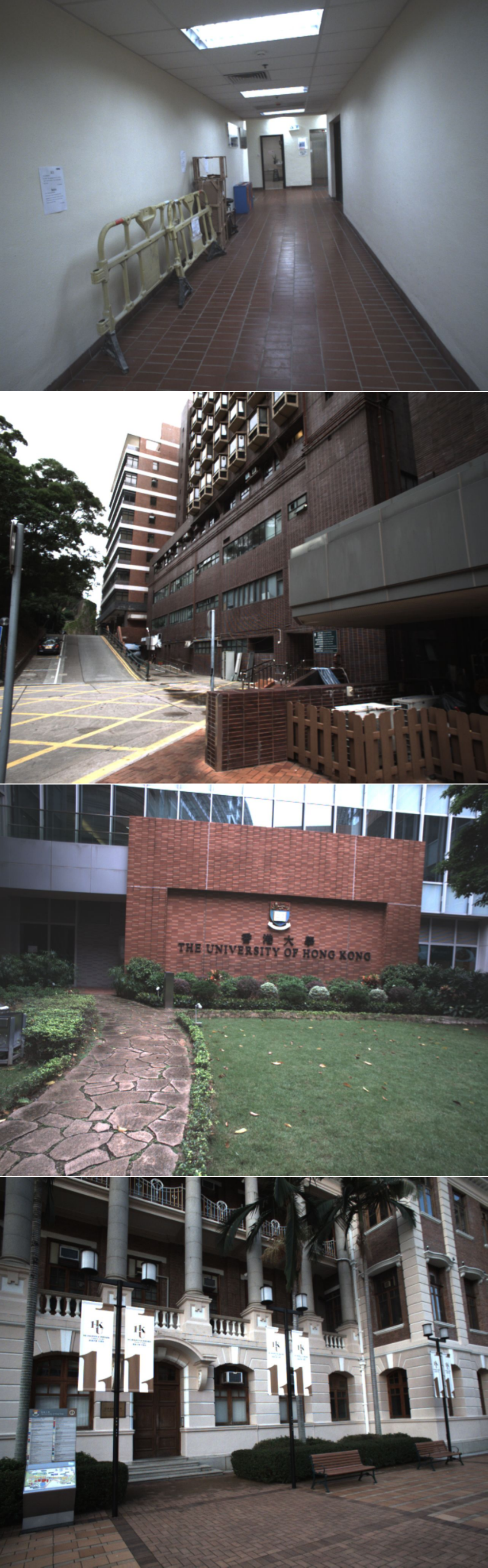}
        \caption{Groundtruth}
        \label{fig:qualitativecomparisongroundtruth}
    \end{subfigure}
    \caption{Qualitative performance comparison of diverse 3DGS SLAM systems.}
    \label{fig:qualitativecomparison}
    \vspace{-1.5em}
\end{figure*}

\subsection{Gaussian Mapping}
Upon receiving each keyframe, we initialize 3D Gaussians. For the first frame, we process the entire point cloud, extracting the 3D coordinates of the points as the centres of the 3D Gaussians. We calculate the squared Euclidean distance of each point to the origin, ensuring a minimum value to prevent zero distances. Opacity parameters are initialized using an inverse Sigmoid function. For colour information, we initialize a tensor to store features extracted from the point cloud's colour data, with RGB channels corresponding to spherical harmonics coefficients. Although we employ spherical harmonics (SH), the initial SH degree is set to 0. As optimization iterations and keyframes increase, the SH degree gradually increases to better fit multiple perspectives, capped at 3.

We optimize each received keyframe once as a submap. Subsequently, in managing the keyframe sequence, upon receiving a new frame, we shuffle all keyframes randomly and then select one frame for optimization at random. To ensure consistency in optimizing each keyframe and maintaining map integrity, we assign an upper limit to the number of optimization iterations for each keyframe. Keyframes that reach this limit are removed from the keyframe sequence.

We optimize the parameters of 3D Gaussians, including rotation, scaling, density, and spherical harmonic coefficients (SH), by minimizing the image loss $\mathcal{L}_c$ and geometric loss $\mathcal{L}_d$.

\begin{equation}
\mathcal{L}={\mathcal{L}_c} + {{\lambda}_d}{\mathcal{L}_d}
\label{eq:mathcal{L}}
\end{equation}

The image loss is composed of the illuminance error and the Structural Similarity between two images (SSIM) error:

\begin{equation}
\mathcal{L}_c=(1-\lambda){\mathcal{L}_1}(I,\mathcal{C}(\mathcal{G},{T_c})) + \lambda{\mathcal{L}_{ssim}}
\label{eq:mathcalLc}
\end{equation}

and geometric loss is defined as the $\mathcal{L}_1$ loss between rendered depth $\mathcal{D}$ and the LiDAR measurement’s depth $\mathcal{D}_{lidar}$:

\begin{equation}
\mathcal{L}_d = \sum \left\lVert \mathcal{D} - \mathcal{D}_{lidar} \right\rVert
\label{eq:mathcalLd}
\end{equation}

\section{EXPERIMENTS SETUP}

\begin{figure*}[ht]  
        \captionsetup{justification=justified, labelsep=colon}
        \centering
        \includegraphics[width=2.0\columnwidth]{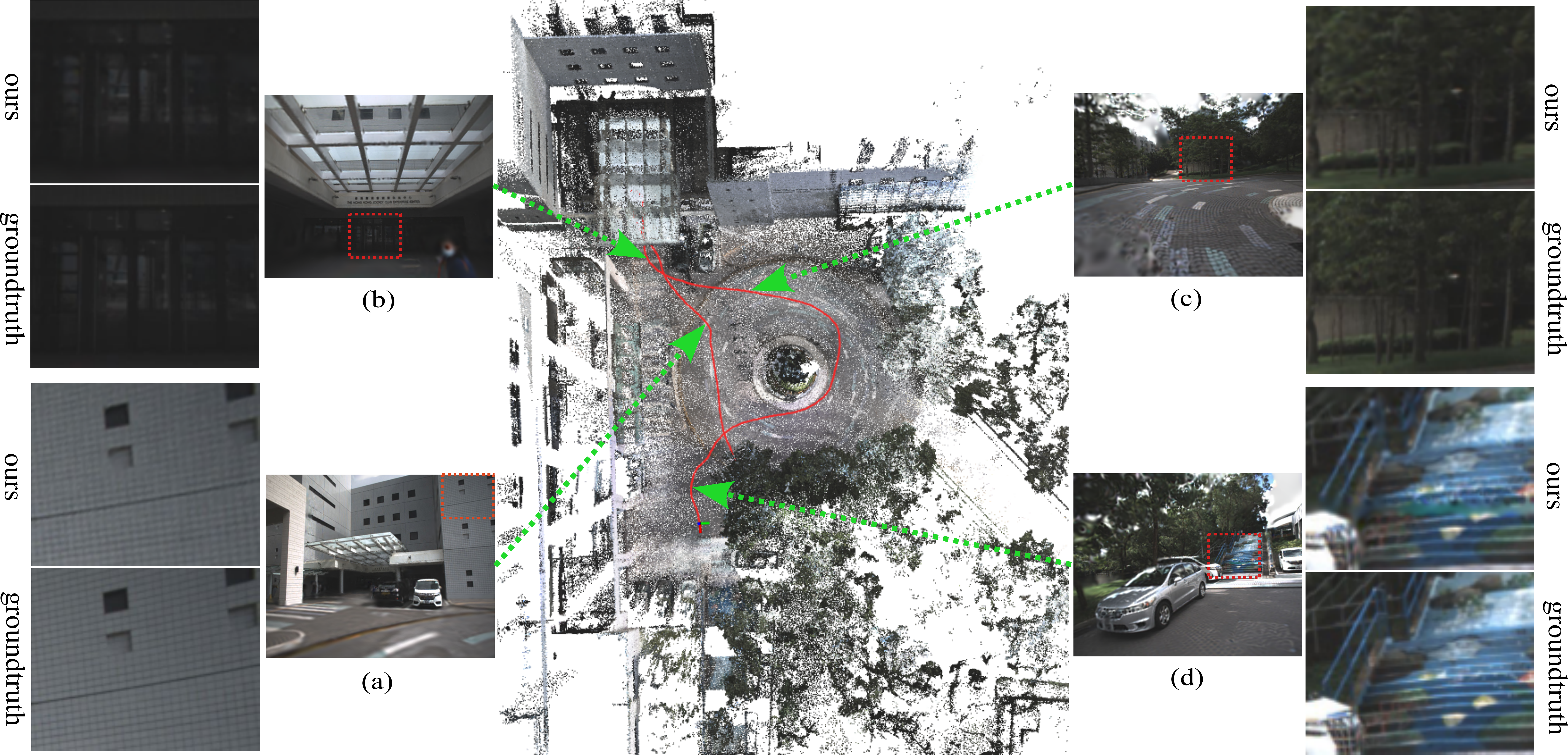}
        \caption{Rendering examples and detailed views of four frames from the hkust_campus_00 (m2) sequence~\cite{r3live}. The red line represents the running trajectory, while (a)-(d) show four selected rendered images at various positions, highlighting details such as glass surfaces, tree branches, and steps. The area within the red rectangle is enlarged to facilitate comparison with the groundtruth.}
        \label{fig:fullseq}
\end{figure*}%

\subsection{Datasets}
We conducted extensive experiments on nine datasets containing LiDAR-Visual-Inertial data. These include hku2(f0), LiDAR_Degenerate(f1), and Visual_Challenge(f2), which are derived from the FAST-LIVO\cite{fastlivo} dataset, captured using two industrial cameras and a Livox Avia LiDAR. Additional datasets include hku_campus_seq_00(r0), degenerate_seq_00(r1), degenerate_seq_01(r2), degenerate_seq_02(m0), hku_campus_seq_01(m1), and hkust_campus_00(m2) from the R3LIVE\cite{r3live} dataset, captured using a global shutter camera and a Livox Avia LiDAR. In particular, the m1, m2, and m3 sequences are truncated to the first 100 seconds to serve as the test datasets. Among these, f1 and m0 represent indoor scenes, while the remaining datasets are outdoor scenes. All images were set to a resolution of 640 × 512.

\subsection{Implementation Details}
We run our system on a desktop with Intel Core i7 13700K 2.50 GHz, a single NVIDIA GeForce RTX 4070 Ti and 32 GB RAM. All of our code is written in C++, with the 3D Gaussian components relying on the LibTorch framework. Time-critical rasterization and gradient computations are executed using CUDA.

\subsection{Baseline Methods} 
We first conducted a comparative analysis of several existing SLAM methods based on NeRF and Gaussian Splatting techniques, including NeRF-SLAM\cite{nerfslam}, MonoGS\cite{monogs}, SplaTAM\cite{splatam}, and Gaussian-LIC\cite{gaussianlic}. Additionally, we evaluated popular open-source RGB- or RGB-D-based SLAM methods, such as MonoGS\cite{monogs} and Photo-SLAM\cite{photoslam}, as well as RGB-D-based methods, including SplaTAM\cite{splatam} and Gaussian-SLAM\cite{gaussianslam}, on a new dataset. Since our dataset lacks depth camera data, we projected LiDAR point clouds onto the pixel plane for each frame to construct sparse pseudo-depth maps, which were used for evaluating the RGB-D-based algorithms. Furthermore, most of the methods (except Photo-SLAM) experienced bad estimations in the tracking modules. To address this, we used our tracking module's pose estimation to replace the pose input of these methods and focused solely on comparing the mapping performance.

\subsection{Metrics}
\label{subsec:metrics}
Quantitative measurements in terms of Peak Signal Noise Ratio(PSNR), Structural Similarity (SSIM), and Learned Perceptual Image Patch Similarity (LPIPS) are adopted to analyze the performance of photorealistic mapping. We also report the computing resources' requirements by showing the optimization time.

\section{EVALUATION}
\subsection{Quantitative Evaluation}
\label{subsec:quantitativeevaluation}
To evaluate mapping performance, we base our assessment on rendering RGB images generated from the constructed maps. The specific metrics used for evaluation are detailed in Section \ref{subsec:metrics}. Table \ref{tab:quantitativecomparison1} summarizes the quantitative performance of the tested methods. Some results are derived from \cite{gaussianlic}. From these results, it is evident that NeRF-SLAM \cite{nerfslam}, despite achieving acceptable performance by incorporating additional depth information from DroidSLAM \cite{droidslam}, continues to focus on generating full-resolution images using neural implicit representations. In contrast, SplaTAM \cite{splatam} emphasizes faster execution by using isotropic 3D Gaussians to model scenes, intentionally disregarding view-dependent effects. While this optimization significantly enhances processing speed, it compromises visual quality and leads to performance degradation in complex, unbounded environments. 

Table \ref{tab:quantitativecomparison2} compares several open-source RGB- and RGB-D-based methods across various datasets. Our experiments confirm similar trends: For RGB-D methods, such as SplaTAM \cite{splatam} and Gaussian-SLAM \cite{gaussianslam}, the initialization of 3D Gaussians relies heavily on depth maps. The inherent sparsity of pseudo-depth maps derived from LiDAR point clouds results in inaccurate Gaussian initialization, consequently leading to suboptimal mapping performance. MonoGS performs effectively in slower-moving indoor scenes but exhibits reduced mapping quality as the scene size increases or motion speed escalates. Photo-SLAM \cite{photoslam}, leveraging ORB-SLAM3 \cite{orbslam3} for continuous feature point initialization, demonstrates relatively better metrics than other RGB- or RGB-D-based 3DGS SLAM methods. Our algorithm achieves the best rendering results compared to the aforementioned methods.

\begin{figure}[h]
    \centering
    \includegraphics[width=0.9\columnwidth]{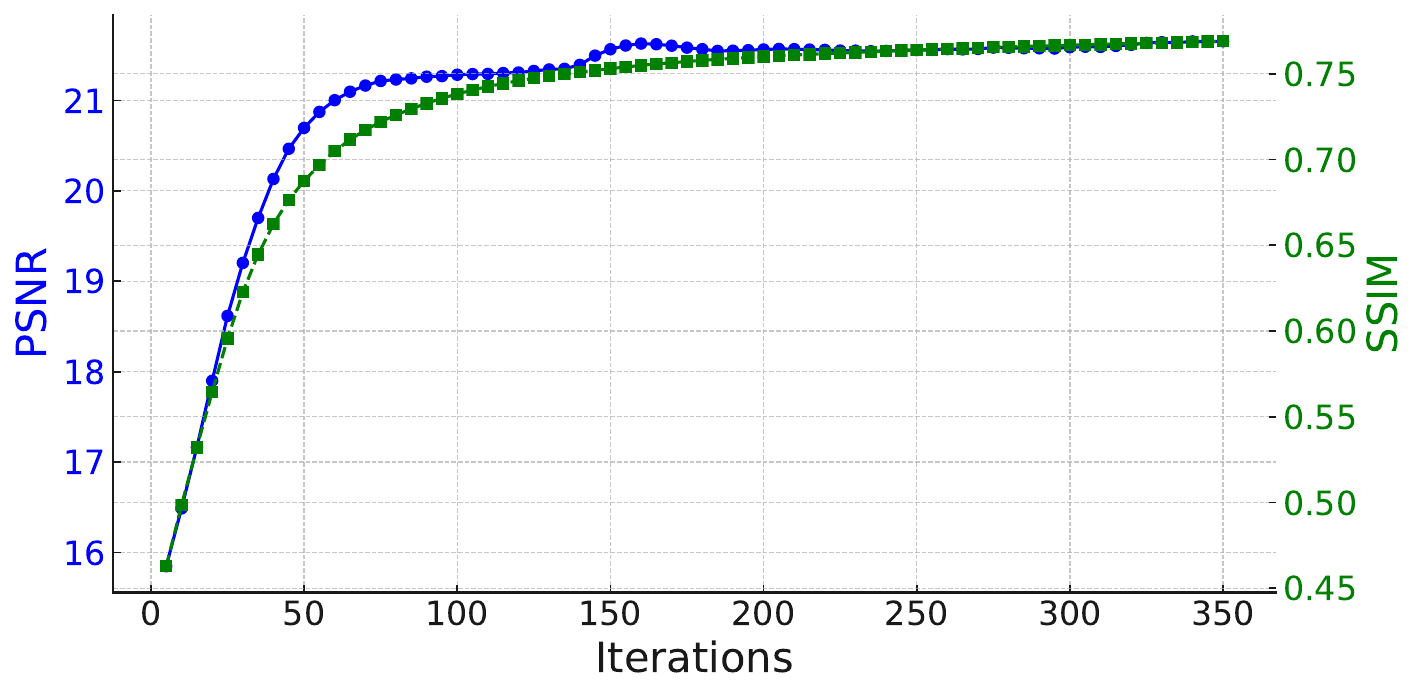}
    \caption{Training metrics (PSNR and SSIM) over iterations for a keyframe.}
    \label{fig:psnrssim}
    \vspace{-2.0em}
\end{figure}

\begin{figure}[h]
    \begin{subfigure}{0.15\textwidth}
        \includegraphics[width=\linewidth]{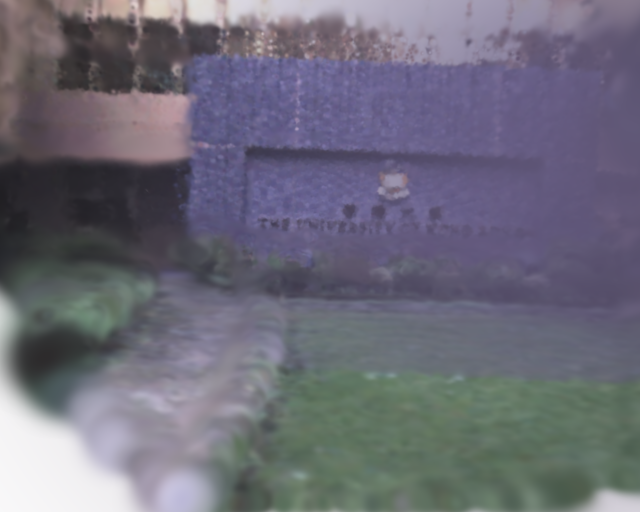}
        \caption{iteration = 5}
        \label{fig:runtime5}
    \end{subfigure}
    \hspace{0.0001\textwidth} 
    \begin{subfigure}{0.15\textwidth}
        \includegraphics[width=\linewidth]{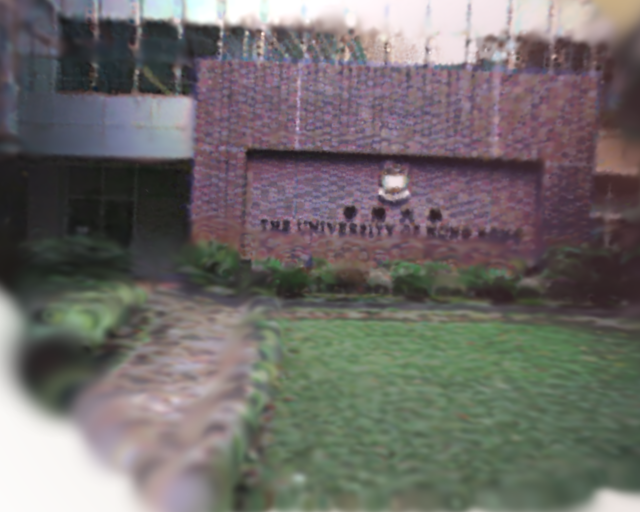}
        \caption{iteration = 30}
        \label{fig:runtime30}
    \end{subfigure}
    \hspace{0.0001\textwidth} 
    \begin{subfigure}{0.15\textwidth}
        \includegraphics[width=\linewidth]{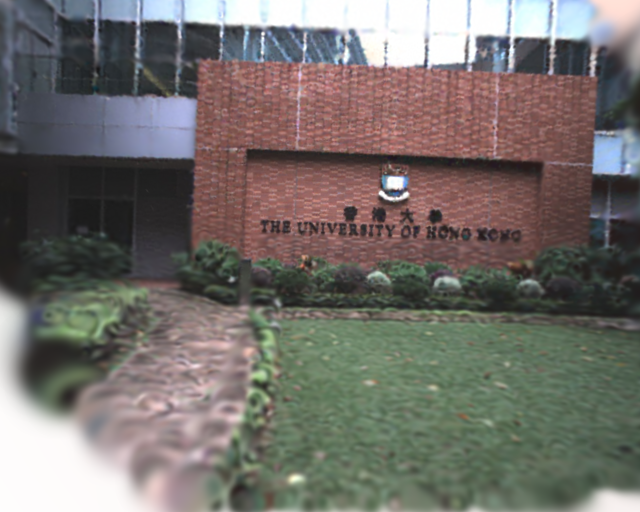}
        \caption{iteration = 105}
        \label{fig:runtime105}
    \end{subfigure}
    \hspace{0.0001\textwidth} 
    \caption{Rendering images at different training iterations.}
    \label{fig:runtime}
    \vspace{-2.0em}
\end{figure}

In our runtime analysis, we tracked PSNR and SSIM for a specific keyframe at each iteration, as shown in Figure \ref{fig:psnrssim}. As illustrated in Figure \ref{fig:runtime}, photo-realistic rendering quality was achieved after approximately 105 iterations, with a total runtime of around 3 seconds.

\begin{table*}[htbp]
    \centering
    \renewcommand{\arraystretch}{1.0} 
    \setlength{\tabcolsep}{10pt} 
    \caption{Quantitative performance comparison of RGB- or RGB-D-based 3DGS SLAM systems on sequences degenerate_seq_02(m0), hku_campus_seq_01(m1), and hkust_campus_00(m2).}
    \label{tab:quantitativecomparison2}
    \begin{tabular}{@{}lccccc@{}}
        \toprule
        \multirow{2}{*}{\textbf{Metrics}} & \multicolumn{5}{c}{\textbf{Method}} \\
        \cmidrule(lr){2-6}
        & MonoGS\cite{monogs} & SplaTAM\cite{splatam} & Gaussian-SLAM\cite{gaussianslam} & Photo-SLAM\cite{photoslam} & Our LVI-GS \\
        \midrule
        PSNR$\uparrow$ (m0) & 21.90 & 12.78 & 6.14 & 27.77 & \textbf{28.99} \\
        PSNR$\uparrow$ (m1) & 11.19 & 14.03 & 9.23 & 19.41 & \textbf{20.17} \\
        PSNR$\uparrow$ (m2) & 15.94 & 9.81 & 8.33 & 21.16 & \textbf{23.93} \\
        \midrule
        \textbf{PSNR$\uparrow$ Avg.} & 16.34 & 12.21 & 7.90 & 22.78 & \textbf{24.36} \\
        \midrule
        SSIM$\uparrow$ (m0) & 0.803 & 0.254 & 0.265 & 0.889 & \textbf{0.899} \\
        SSIM$\uparrow$ (m1) & 0.231 & 0.467 & 0.349 & 0.658 & \textbf{0.728} \\
        SSIM$\uparrow$ (m2) & 0.532 & 0.244 & 0.274 & 0.698 & \textbf{0.784} \\
        \midrule
        \textbf{SSIM$\uparrow$ Avg.} & 0.522 & 0.322 & 0.296 & 0.748 & \textbf{0.804} \\
        \midrule
        LPIPS$\downarrow$ (m0) & 0.804 & 0.860 & 0.907 & 0.311 & \textbf{0.173} \\
        LPIPS$\downarrow$ (m1) & 0.231 & 0.632 & 0.869 & 0.271 & \textbf{0.211} \\
        LPIPS$\downarrow$ (m2) & 0.532 & 0.827 & 0.925 & 0.378 & \textbf{0.223} \\
        \midrule
        \textbf{LPIPS$\downarrow$ Avg.} & 0.522 & 0.773 & 0.900 & 0.320 & \textbf{0.202} \\
        \bottomrule
    \end{tabular}
    \captionsetup{justification=centering, labelsep=colon} 
\end{table*}

\subsection{Qualitative Evaluation}
In this section, we present the rendering images generated by the methods tested in Section \ref{subsec:quantitativeevaluation}. Each method was run for the same number of iterations and rendering from identical viewpoints, facilitating a qualitative comparison to examine visual differences among the approaches. Comparisons with RGB-D-based 3DGS SLAM methods were omitted due to their limited rendering quality.

We also compared MonoGS \cite{monogs}, Photo-SLAM \cite{photoslam}, our proposed method, and the ground truth, as illustrated in Figure \ref{fig:qualitativecomparison}. Our observations show that MonoGS \cite{monogs} performs adequately in indoor environments; however, it exhibits noticeable blurring in outdoor scenes. In contrast, our method demonstrates a marked improvement over Photo-SLAM \cite{photoslam} in restoring texture details on surfaces such as floors and walls. By leveraging the denser spatial point clouds generated by LiDAR in texture-rich regions, our method achieves superior detail restoration within the same number of training iterations. Figure \ref{fig:fullseq} provides further examples of rendering details across four frames from sequence hkust_campus_00(m2)\cite{r3live}. Even in challenging cases, such as scenes containing dense textures, glass surfaces, tree branches, and steps, our method consistently maintains high rendering quality.

\subsection{Ablation Study}

Table \ref{tab:ablation} illustrates the impact of depth loss and different pyramid levels on training outcomes. For this ablation study, we selected a segment from sequence m1 that contains abundant point clouds and distant scenes to minimize interference from insufficient point clouds and limited depth variation. Our findings indicate that evaluation metrics improve as the pyramid level increases, reaching an optimal performance at $l=2$. We also conducted comparisons without depth loss and observed a modest improvement in rendering quality with its inclusion.

\begin{figure}[h]
    \begin{subfigure}{0.15\textwidth}
        \includegraphics[width=\linewidth]{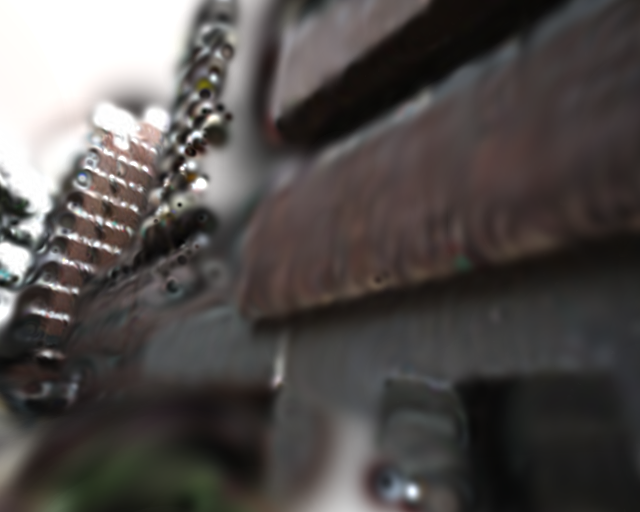}
        \caption{$l=2$ without depth loss}
        \label{fig:ablationwod}
    \end{subfigure}
    \hspace{0.0001\textwidth} 
    \begin{subfigure}{0.15\textwidth}
        \includegraphics[width=\linewidth]{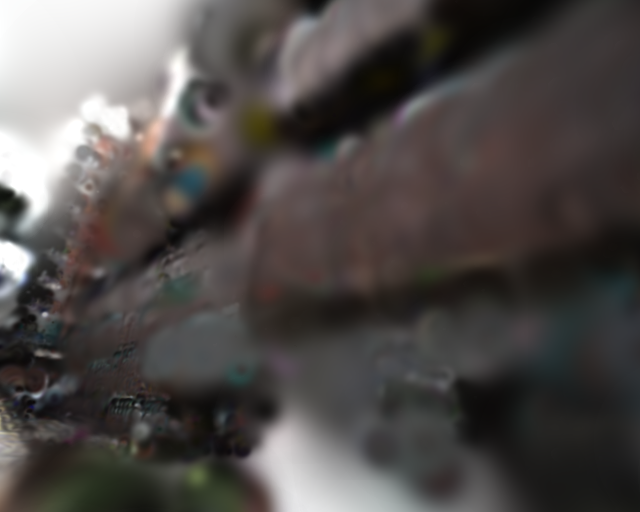}
        \caption{$l=3$ with depth loss}
        \label{fig:ablationwd3}
    \end{subfigure}
    \hspace{0.0001\textwidth} 
    \begin{subfigure}{0.15\textwidth}
        \includegraphics[width=\linewidth]{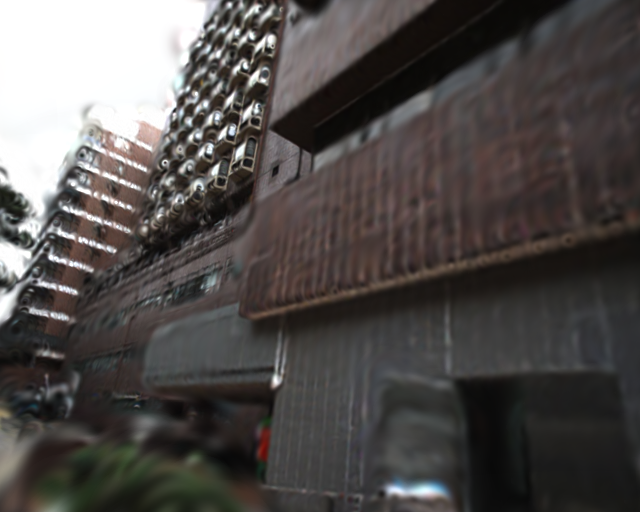}
        \caption{$l=2$ with depth loss}
        \label{fig:ablationwd2}
    \end{subfigure}
    \hspace{0.0001\textwidth} 
    \caption{Rendering results across different control groups.}
    \label{fig:ablation}
\end{figure}

Alongside the quantitative results, Figure \ref{fig:ablation} presents rendering images for visual comparison. When depth loss is omitted, quality degradation is evident in regions with significant depth variation. Additionally, when the pyramid level $l$ is set to 3, lower-resolution pyramid images used in training lead to a substantial loss of scene detail, resulting in insufficiently trained fine-grained features.

\begin{table}[h]
    \centering
    \renewcommand{\arraystretch}{1.0} 
    \setlength{\tabcolsep}{9pt} 
    \caption{Ablation study on the effect of depth loss and pyramids-based training.}
    \label{tab:ablation}
    \begin{tabular}{@{}cccccc@{}}
        \toprule
        \multicolumn{3}{c}{\textbf{Control Groups}} & \multicolumn{3}{c}{\textbf{Metrics}} \\
        \cmidrule(lr){1-3} \cmidrule(lr){4-6}
        \# & Depth Loss & Pyramids & PSNR $\uparrow$ & SSIM $\uparrow$ & LPIPS $\downarrow$ \\
        \midrule
        (1) & w/ & w/o & 20.40 & 0.706 & 0.318 \\
        (2) & w/ & $l=1$ & 21.77 & 0.732 & 0.271 \\
        (3) & w/ & $l=2$ & \textbf{22.36} & \textbf{0.747} & \textbf{0.269} \\
        (4) & w/ & $l=3$ & 18.91 & 0.632 & 0.475 \\
        (5) & w/o & $l=2$ & 21.55 & 0.739 & 0.302 \\
        \bottomrule
    \end{tabular}
    \captionsetup{justification=centering, labelsep=colon} 
\end{table}

\section{CONCLUSION}

In this paper, we presented LVI-GS, a tightly-coupled LiDAR-Visual-Inertial SLAM system utilizing 3D Gaussian Splatting (3DGS) for real-time, high-fidelity scene reconstruction and rendering. Our approach leverages both LiDAR and image data, enabling the capture of precise geometric structures and detailed visual information, even in challenging outdoor environments. Our system achieves photorealistic mapping quality with notable computational efficiency through the effective integration of Gaussian map expansion, keyframe management, thread management and CUDA-based acceleration.

Extensive experiments demonstrate that LVI-GS outperforms existing state-of-the-art RGB- or RGB-D-based 3DGS SLAM systems, particularly in maintaining rendering quality and efficiency across various complex scenarios. Our ablation studies further validate the benefits of pyramid-based training and depth loss for enhancing map representation accuracy. Future work will explore the integration of additional sensor modalities and further optimization of the framework for broader applicability in real-time robotic applications and AR/VR environments.


\bibliographystyle{IEEEtran}
\bibliography{LVI-GS}

\end{document}